# Suspicious Vehicle Detection Using Licence Plate Detection And Facial Feature Recognition

Vrinda Agarwal, Aaron George Pichappa, Manideep Ramisetty, Bala Murugan MS, Manoj kumar Rajagopal

## ABSTRACT

With the increasing need to strengthen vehicle safety and detection, the availability of pre-existing methods of catching criminals and identifying vehicles manually through the various traffic surveillance cameras is not only time-consuming but also inefficient. With the advancement of technology in every field the use of real-time traffic surveillance models will help facilitate an easy approach.
Keeping this in mind, the main focus of our paper is to develop a combined face recognition and number plate recognition model to ensure vehicle safety and real-time tracking of running-away criminals and stolen vehicles.

## INTRODUCTION

In the past few decades, there has been an expanding number of vehicles on roads. Vehicles are increasingly used in organized crimes and often as a means of transport to flee from criminal activities. With the increasing connectivity of roads, it is essential to detect vehicles in real-time to manage the traffic, identify stolen/blacklisted vehicles, and control transport access to these vehicles. As a result, several efforts are being made to help solve this problem.
There are two significant elements when talking about vehicle detection and tracking and criminal recognition, the first being able to track vehicles using their license recognition and the other being able to recognize the person's face in that vehicle.
The reason to combine the facial recognition concept with license plate recognition is that criminals are known to either ditch their vehicle for a new one or change their license plate in order to lose their trail, that's when facial recognition will help us detect the new vehicle they are traveling in.

We have two significant models associated with our paper: facial feature recognition and number plate recognition, which are discussed at length in this paper.

The rest of the paper is organized as follows. Section II briefs the literature of previous works done in the field. Section III details the proposed methodology. Section IV discusses the experimental results of the datasets and the analysis of the models. Finally, we summarized the conclusion in Section V and future work in Section VI.

## II. LITERATURE REVIEW

In various existing literature, authors with work related to surveillance for security focus specifically on facial feature recognition, they are using varied approaches such as:

[2] Yougjin Lee, Kyunghee Lee and Sungbum Pan, *"Local and Global Feature Extraction for Face Recognition"*, LFA is a method that constructs kernels to detect the local structures of a face but this, however, addresses only image recognition the paper talks about the problematic features of LFA and alternatively proposes a new feature extraction method

[4] Hanaa Mohsin Ahmed and Rana Talib Rasheed, *"A Raspberry PI Real-Time Identification System on Face Recognition"*, Raspberry-Pi-based systems to upgrade face recognition to a level that it substitutes the use of passwords and RFI-Cards for system security

[8] Mila Mileva and A. Mike Burton, *"Face Search in CCTV Surveillance"* talks about face search in CCTV, which is essentially a visual search experiment. Viewers are expected to find faces in CCTV surveillance and identify faces, they are given face references by using images from the respective person's photo IDs and passports. Results talk about performance gain in recognition with the use of more target images.

[9] A paper titled *"Video Surveillance Framework based on Real-Time Face Mask Detection and Recognition"*, proposes real-time face detection and recognition using CCTV surveillance camera videos. The proposed system is divided into various phases: video acquisition, where keyframes are chosen using the HoG algorithm followed by keyframes selection, data augmentation includes CLAHE with normalization using Angular Affine Transformation, feature division following pixel-based feature extraction done by clustering approach called (EM-GMM) and Yolo Nano technique respectively, the features that were extracted were used in Hassanat similarity with KNN to build a codebook, then after mask detection the final stage being face detection.

Talking about works with respect to traffic surveillance, authors have proposed different systems:

[7] Varsha Sahadev Nagmode and S.M. Rajbhoj, *"An IOT platform for vehicle traffic monitoring system and Controlling System based on priority"* focuses on IoT and sensing technologies, the system uses various sensors the major sensor is ultrasonic sensors. It helps in knowing the traffic level, and the controlled data recorded is sent to the web server through a wifi module, which is another major sensor. Based on the data and collected the traffic lights are manipulated such that the highest priority is given to lanes with high-level traffic. This communication in the main system is done using RF transceivers. This system is used at lane intersections and has high reliability.

[11] Tanushri Bhagat, and Rahul Thakur, *"Automatic Recognition of License Plates"*, acronym ALPR uses cutting-edge image processing via Python packages. The paper is primarily talking

abut the effective use of this system in India and connects this to the national databases: VAHAN and SARTHI. It also adds the significance of detecting high-intensity images in India.

[13] Qi Wang, Xiaocheng Lu, Cong Zhang, Yuan Yuan, and Xuelong Li, *"LSV-LP: large-scale video-based license plate detection and recognition"* uses (CNN) for LPDR systems. The authors have created an LSV-LP dataset, made up of annotated license plates made up of a large number of films, frames, and videos. The efficiency of the proposed system is confirmed by the ablation studies, also the rigorous tests comparing it with the existing model show appreciable results.

[12] Cong Zhang, Qi wang, and Xuelong Li in their paper talk about focusing on video contextual information, instead of the existing systems which are based on a single-image-based algorithm for license plate detection and also don't perform well in unrestricted circumstances with moving objects. The existing image-based models have restrictions in terms of distorted images and videos taken in an intricate environment. They have used a deep-learning-based approach that is capable of automatically detecting, tracking, and recognizing license plates in dynamic and unconstrained, and complex scenarios. Each of the components has great efficiency which is combined to evaluate the proposed technique.

[5] A paper published in IEEE CCDC in 2018 proposes a vehicle tracking system that fuses the radar and V2V information to improve target tracking accuracy. The system integrates DSRC, Dedicated short-range communication, GPS, Global Positioning System and radar. Karman filter is used to improve the accuracy of tracking the target. Along with the proposed system the paper also talks about resolving the problem of data association in multiple target tracking measurements by connected vehicle information.

[19] The paper presented at the AIP conference proposes recognition of license plates using a key-point search such that they can be conducted using an indirect method of convolutional regression. The system is using CNN with convolutional and de-convolution layers to enable it to perform regression. This is also evaluated on video frames collected from surveillance cameras and the results achieved using CNN are better than the Faster R-CNN.

As stated above we can conclude that there are various works and measures have been taken, for building a better surveillance system but each of them addresses one single objective, either license plate recognition or face recognition. There however have been a few works combining the above two ideas for better surveillance of suspicious vehicle detection.

[21] Fikret Alim, Enes Kavakli, Sefa Burak Okcu, Ertugrul Dogan and Cevahir Cgila, *"Simultaneous License Plate Recognition and Face Detection at the Edge".* The authors of the paper talk about using low-power Neural Processing Units (NPU) inside embedded Systems on Chips(SoC) and advances in small single-board computers with high parallel processing power to enable real-time face detection and License Plate Detection (LPR) at the edge. This poses a problem as running multiple algorithms concurrently with high accuracy and prompt execution remains a challenge.

With our paper, we are combining both these objectives of face recognition and license plate detection with optimizing AI models to suit our needs and thus strengthen the vehicle detection system, in real-time with effectivity.

## III. PROPOSED METHODOLOGY

We divide the proposed methodology into three phases: first being face recognition, second being license plate detection, and third linking the two models to run simultaneously.

*FACE RECOGNITION:*

We use open CV and Python programming language for face recognition. For the pre-trained model, we are using pre-installed Python libraries, namely 'face_recognition' and 'face_recognition_models'. Then we create two separate self-made databases, one containing known faces which are essentially just face shots, and the other with unknown faces with images of multiple objects and faces. We are using CNN as a model for face detection. We are iterating the known face over the unknown face images and generating an array with boolean results of match (true) and unmatched (false). Then we extract the image where the known face matches.

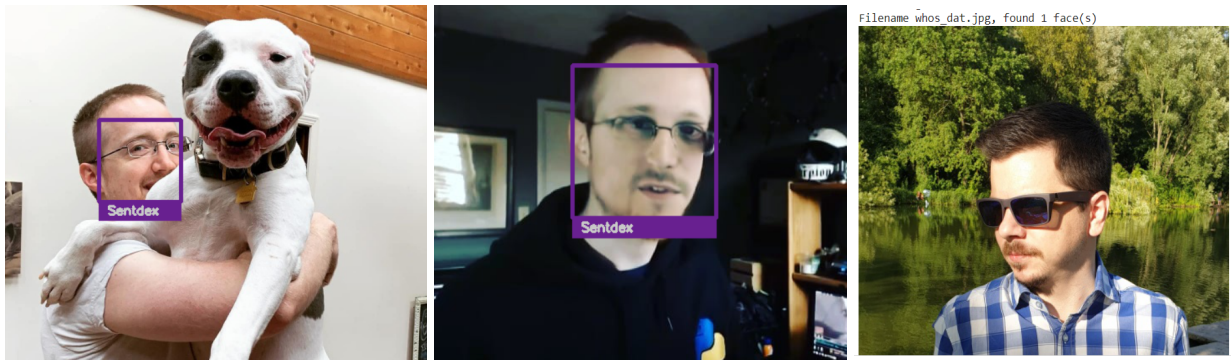

*Figure 3.1. Face recognition*

*LICENSE PLATE DETECTION:*

For License Plate Detection we are using Google Colab and Python programming. We are importing and installing the required repositories. The model that we are using is YOLOv8, which is a new-age model for object detection, image classification, and instance segmentation tasks. The dataset that we are using for training the model is 'roboflow'. After the detection of the license plate, we implement the code to be able to read the license plate number. We are using an external video for testing the model. The model correctly reads the number on the license plate and displays it on the video as shown below.

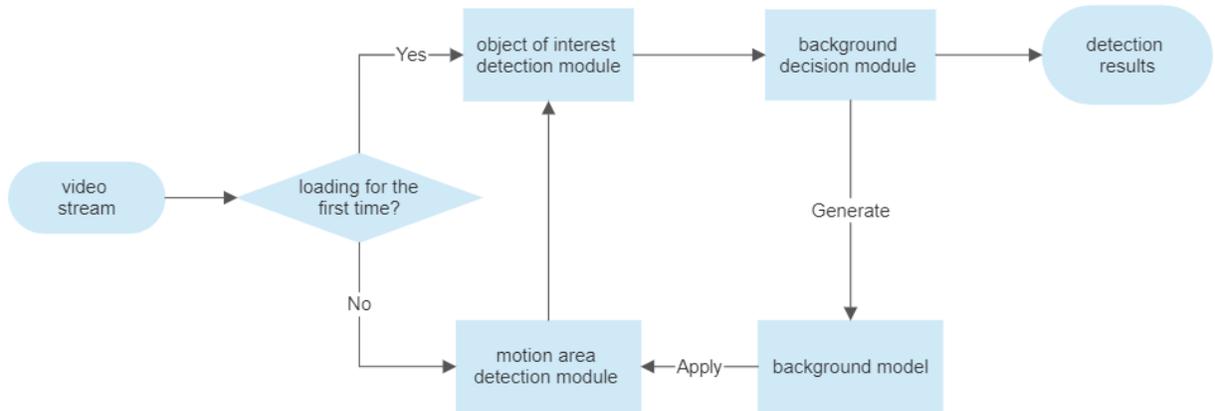
*Flowchart 1. license plate recognition*

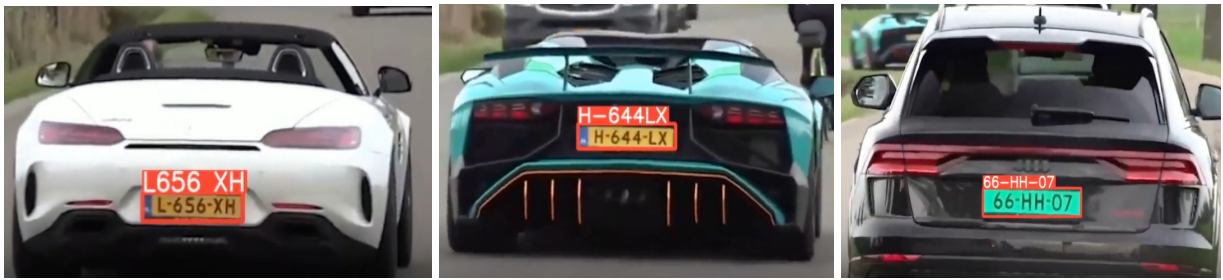
*Figure 3.2. license plate recognition*

For phase three of the paper, we have combined the two models to work simultaneously in a single place to enable their work hand-in-hand in order to achieve the objective of the proposed system.

## IV. RESULTS

The figures below show the confusion matrix (which helps us in summarising the model visually to depict model performance) for the model that we developed for License Plate Recognition, which shows an accuracy of 87%. And the other image on the right depicts the training and validation losses among which our main concern is the box-loss (representing how well the algorithm can locate the center of the license plate) and cls-loss (which measures the classification of each bounding box) are depicted inside the yellow marked graphs respectively.

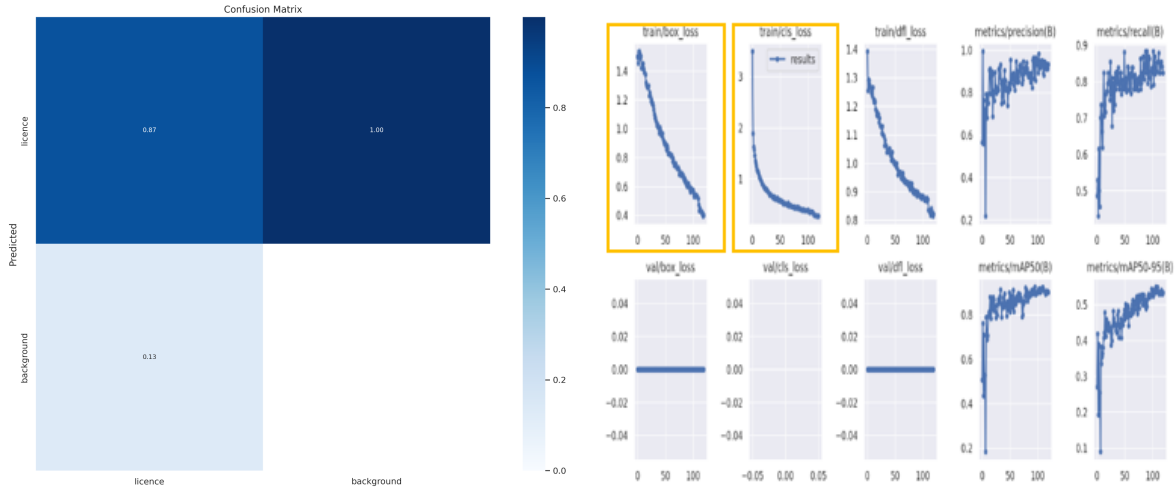

*Figure 4.1. the right image is a confusion matrix and the left one is the validation losses*

From the model summary table, (figure 4.2) we can see that we are getting a mean average precision (mAP50) of 0.926.

| Class | Images | Instances | Box (P | R | mAP50 | mAP50-95) |
|---|---|---|---|---|---|---|
| all | 64 | 68 | 0.961 | 0.838 | 0.926 | 0.582 |

*Figure 4.2 (a). Model Summary for LPD*

```
Model summary: 218 layers, 25840339 parameters, 0 gradients, 78.7 GFLOPs
val: Scanning /content/Automatic_Number_Plate_Detection_Recognition_YOLOv8/ultralytics/yolo/v8/detect/Car_License_Plates-1/valid/labels.cache... 64 images,
       Class     Images  Instances      Box(P          R      mAP50  mAP50-95): 100% 4/4 [00:03<00:00,  1.09it/s]
         all         64         68      0.961      0.838      0.926      0.582
Speed: 3.0ms pre-process, 24.8ms inference, 0.0ms loss, 2.5ms post-process per image
```

*Figure 4.2 (b). Model Summary for LPD*

For the face recognition model, the accuracy of the pre-trained model is 99.38% which is further tested on a custom dataset.

## V. CONCLUSION

In this paper, we combined face recognition and license plate detection and recognition, in order to achieve the need to strengthen vehicle safety and detection. The proposed system is a real-time traffic surveillance model which is an easy approach against the traditional inefficient and manual ways of surveillance. The proposed system is able to run at high performance and accuracy. The presented examples indicate that the fusion of the two models with their outputs and high accuracy will be very helpful in security and surveillance.